\pdfoutput=1

\documentclass[11pt]{article}

\usepackage[final]{acl}
\usepackage{times}
\usepackage{latexsym}
\usepackage[T1]{fontenc}
\usepackage[utf8]{inputenc}
\usepackage{microtype}
\usepackage{inconsolata}
\usepackage{graphicx}
\usepackage{newfloat}
\usepackage{listings}
\usepackage{float}

\DeclareFloatingEnvironment[%
    listname={List of Prompts},
    name=Prompt,
    placement=tbh,
    within=section
]{prompt}

\title{What does Kiki look like? Cross-modal associations between speech sounds and visual shapes in vision-and-language models}

\author{Tessa Verhoef$^{\ast}$, Kiana Shahrasbi, Tom Kouwenhoven\thanks{Equal contribution} \\
        Leiden Institute of Advanced Computer Science \\ Leiden University,
The Netherlands \\ \texttt{t.verhoef@liacs.leidenuniv.nl, k.shahrasbi@umail.leidenuniv.nl,} \\\texttt{t.kouwenhoven@liacs.leidenuniv.nl}}

\begin{document}
\maketitle
\begin{abstract}
Humans have clear cross-modal preferences when matching certain novel words to visual shapes. Evidence suggests that these preferences play a prominent role in our linguistic processing, language learning, and the origins of signal-meaning mappings. With the rise of multimodal models in AI, such as vision-and-language (VLM) models, it becomes increasingly important to uncover the kinds of visio-linguistic associations these models encode and whether they align with human representations. Informed by experiments with humans, we probe and compare four VLMs for a well-known human cross-modal preference, the bouba-kiki effect. We do not find conclusive evidence for this effect but suggest that results may depend on features of the models, such as architecture design, model size, and training details. Our findings inform discussions on the origins of the bouba-kiki effect in human cognition and future developments of VLMs that align well with human cross-modal associations.
\end{abstract}

\section{Introduction}

The development of machine understanding and generation of natural language has benefited immensely from the introduction of transformer-based architectures \cite{Vaswani2017Attention}. These architectures have since then been adapted and extended to handle multimodal data, leading to the creation of various types of multimodal models, including vision-and-language models. These models can potentially revolutionize how AI systems understand the world and interact with humans. However, we lack direct access to the exact representations and associations they encode. How VLMs integrate representations in the two modalities and whether associations between modalities are made in a human-like way is still being actively investigated \cite{alper2023bert, kamath-etal-2023-whats, zhang2024cross, karamcheti2024prismatic}.

\begin{figure}[t]
  \includegraphics[width=0.48\linewidth]{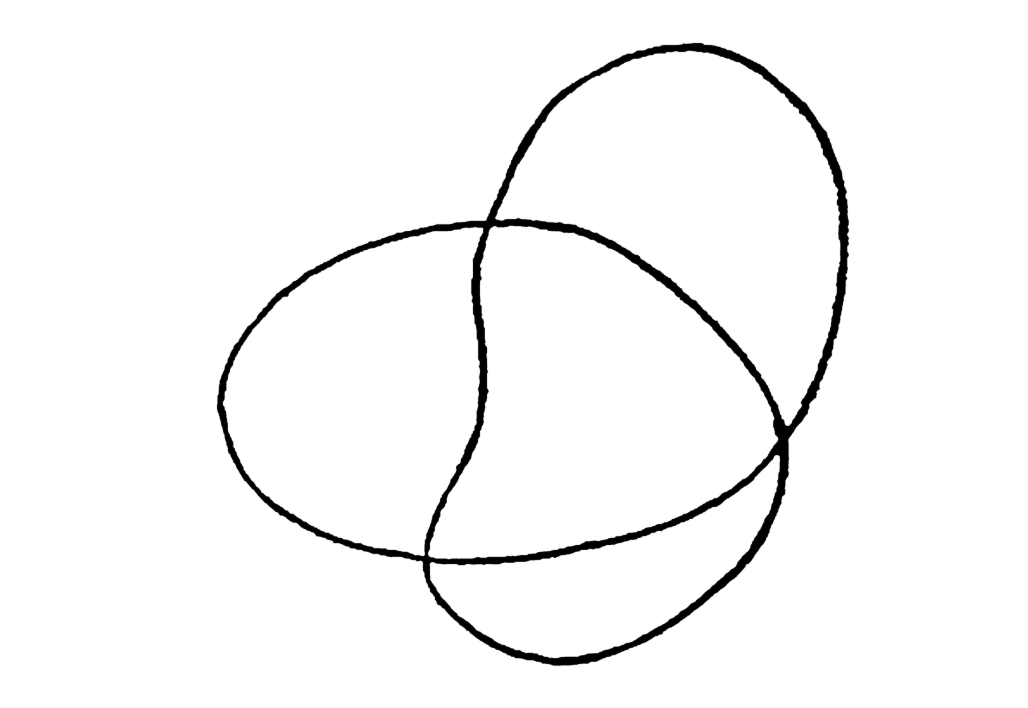}  \hfill
  \includegraphics[width=0.48\linewidth]{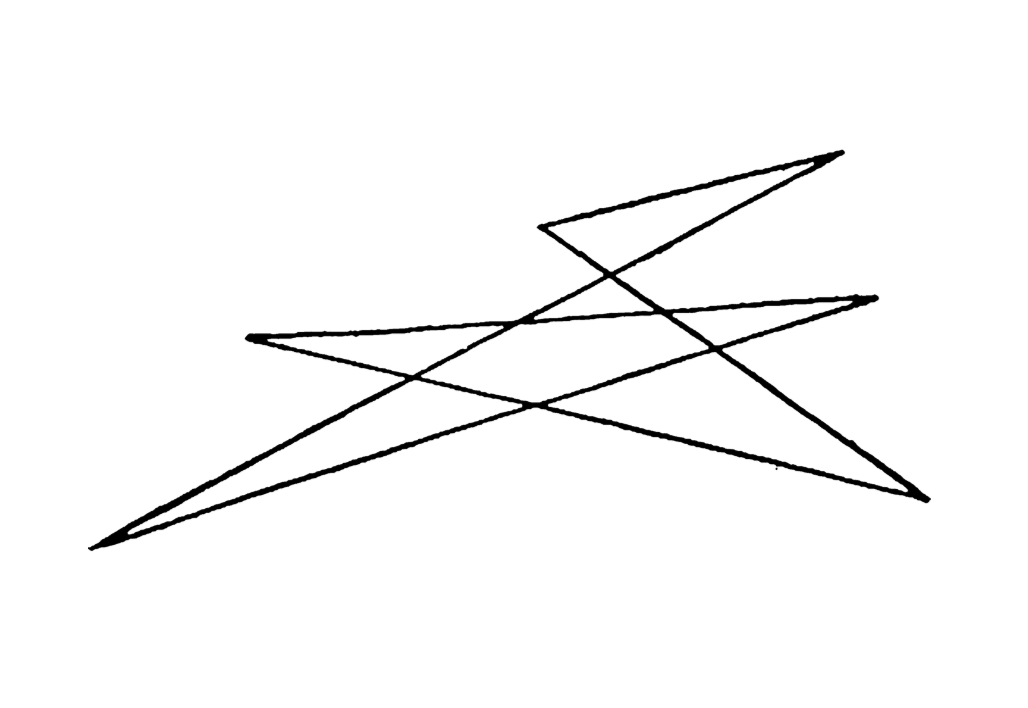}
  \caption{Which of these two shapes is Kiki?  Images from \citet{kohler1929gestalt,kohler1947gestalt}}
  \label{fig:kohler}
\end{figure}

Here, we use a well-known paradigm from the field of cognitive science to probe into a specific cross-modal association between speech sounds and visual shapes: the bouba-kiki effect. When humans see two figures, one with jagged and one with smooth edges, and are told one is a Kiki and the other a Bouba, 95\% will name the jagged figure Kiki \cite{ramachandran2001synaesthesia}. This effect was initially discovered and described anecdotally by Wolfgang K\"ohler \cite{kohler1929gestalt,kohler1947gestalt}, using the two images shown in Figure \ref{fig:kohler} with the labels \textit{maluma} and \textit{takete}. Since then it has been widely studied (as reviewed in Section \ref{bg}), and expanded with many other cross-modal preferences in human processing of (speech) sounds and visual imagery. Moreover, a wealth of evidence suggests that such preferences widely influence patterns we see in human languages \cite[e.g.,][]{ramachandran2001synaesthesia, cuskley2013syn, imai2014sound,verhoef2015emergence, verhoef2016iconicity, tamariz2018interactive}. Even though non-arbitrariness in language is often still regarded as an exception in some disciplines, in fields such as language evolution, and sign language linguistics, iconic form-meaning mappings are considered omnipresent \cite{perniss2010iconicity}. Given the central role cross-modal preferences play in human visio-linguistic representations and their effects on language, it is pertinent to investigate whether VLMs associate non-words and visual stimuli in a human-like way. 

Examining universal human cross-modal preferences in VLMs can help us gain key insights across disciplines. First, it may reveal whether VLMs process multimodal information in a human-like way and whether similar biases drive their understanding of visual-auditory form-meaning mappings. Overlap in cognitive biases can potentially increase mutual understanding and improve interactions between humans and machines \cite{kouwenhoven2022emerging}. Second, it may help pinpoint what is missing to make VLMs more suitable for realistic simulations of human language emergence. Increasingly, VLMs are used in emergent communication settings, where agents communicate with each other and develop a novel language \cite{bouchacourt-baroni-2018-agents, mahaut2023referential, kouwenhoven2024thecurious}. 
These models are used to improve machine understanding of human language \cite{lazaridou2020emergent, Lowe2020On, steinert-threlkeld2022emergent, zheng2024iterated}, but also to simulate and study human language evolution processes \cite{galke2022emergent, lian-etal-2023-communication}. 
While the influence of cross-modal associations on the emergence of language has been studied extensively in language evolution experiments with humans \citep{verhoef2015emergence, verhoef2016iconicity, tamariz2018interactive, little2017signal}, the phenomenon is still absent from current emergent communication paradigms. Evidently, cognitively plausible VLMs are more suitable for simulating aspects of the evolution of meaning in language. 
Finally, the actual origin of the bouba-kiki effect is still being debated within cognitive science and linguistics, with proposed explanations ranging from attributing it to similarities between shape features and features of either orthography \cite{cuskley2017phonological}, acoustics and articulation \cite{ramachandran2001synaesthesia, maurer2006shape, westbury2005implicit}, affective–semantic properties of human and non-human vocal communication  \cite{nielsen2011sound}, or physical properties relating to audiovisual regularities in the environment \cite{fort2022resolving}. If the bouba-kiki effect can be reproduced in a VLM, it can help reveal the crucial ingredients for this effect, potentially leading to models better aligned with human representations. 

To the best of our knowledge, only one previous paper discussed the bouba-kiki effect in VLMs. \citet{alper2024kiki} tested two models, CLIP \cite{radford2021learning} and Stable Diffusion \cite{rombach2022high}, and reported to find strong evidence for the effect in these models. This is somewhat surprising given the way these models are trained and the absence of relevant data sources such as auditory information and experience with physical object properties. Therefore, we introduce nuance in this discussion and show, contrary to this previous finding, that the bouba-kiki effect does not occur consistently in VLMs, and the presence of this cross-modal preference may depend on the way it is tested and properties like model architecture, attention mechanism, and training details. 
 
\section{Background}\label{bg}

\subsection{Sound-symbolism and cross-modal associations in language and cognition}
\label{cross-modal-associations}
When \citet{hockett1960origin} listed a set of design features deemed essential to natural human language, "arbitrariness" was included. This feature refers to the arbitrary/unmotivated mapping between words and their meanings. 
However, when exploring beyond Indo-European languages, non-arbitrary form-meaning mappings appear to play a significant role in many languages \citep{imai2008sound,perniss2010iconicity,dingemanse2012advances}. Most obviously, perhaps, sign languages are rich in non-arbitrary "iconic" mappings, with articulators that lend themselves particularly well to representing meanings by mimicking, for example, shapes or actions. However, some spoken languages also have specific classes of words where characteristics of the meaning are mimicked or iconically represented in the word. Examples have been identified as "ideophones," "mimetics", or "expressives," and this phenomenon is often called sound-symbolism \citep{imai2008sound,imai2014sound, dingemanse2012advances}. Even in languages not typically considered rich in sound symbolism, such as English and Spanish, vocabulary items from specific lexical categories, like adjectives, are rated high in iconicity as well \cite{perry2015iconicity}. Perhaps the most overwhelming evidence for the widespread importance of sound-symbolism in human languages comes from a study by \citet{blasi2016sound}, who analyzed vocabularies of two-thirds of the world's languages and found evidence for strong associations between speech sounds and particular meanings across geographical locations and linguistic lineages. Consequently, non-arbitrariness is an important property of all languages. 

In addition, human language learning, processing, and evolution are affected by cross-modal associations. 
Sound-symbolic mappings help young children acquire new words \citep{imai2008sound}, and iconic words are learned earlier in child language development \cite{perry2015iconicity}. Furthermore, parents use sound-symbolic words in their infant-directed speech more often than in adult-to-adult conversations \citep{imai2008sound}. In a novel word learning task, participants trained on a mapping congruent with a known cross-modal association performed better than participants in an incongruent condition \citep{nielsen2012source}. Sound-symbolic mappings in language have been connected to cross-modal mappings in the human brain \citep{simner2010sound,ramachandran2001synaesthesia, lockwood2015iconicity} and processing of sound-symbolic words is less affected by aphasia (language-affecting brain damage after left-hemisphere stroke), than arbitrary words \cite{meteyard2015semantics}. It is also argued that universally shared cross-modal biases play an essential role in the evolution of language by bridging the gap between sensory input and meaning by providing a basis for linguistic conventions \cite{ramachandran2001synaesthesia, cuskley2013syn, imai2014sound}. Shared biases can help to create mutual understanding because communicative partners will automatically understand what is meant when a word like "kiki" is used for the first time in a context like the one shown in Figure \ref{fig:kohler}. 

While the bouba-kiki effect may be the most famous example of a universal cross-modal association, many other cognitive biases in cross-modal perception have been reported. For example, non-arbitrary associations exist in human processing between high pitch sounds and light
shades \cite{marks1974associations, melara1989dimensional, ward2006sound}, light shades with rising intonation 
\cite{hubbard1996synesthesia}, graphemes and colours \cite{cuskley2019cross}, vowel height and lightness \cite{cuskley2019cross}, small size and high pitch \cite{evans2010natural, parise2009birds} and vowel openness and visual size \cite{schmidtke2014phonological}. Therefore, the findings presented in this paper only scratch the surface of what is possible in this domain.

\subsection{Testing the bouba-kiki effect in humans}
After its initial discovery, the bouba-kiki effect has been studied increasingly rigorously, extending the initial pair of two images with more possible pairs \cite{maurer2006shape, westbury2005implicit}, and even randomly generated ones to control for biases related to deliberate selection by the researchers \cite{nielsen2011sound,nielsen2013parsing}. In addition, various sets of labels and pseudowords have been contrasted and compared to study the relative importance of vowels versus consonants in the labels \cite{westbury2005implicit, nielsen2011sound,nielsen2013parsing}. The role of orthography, in addition to auditory properties of speech sounds, has also been studied \cite{cuskley2017phonological, bottini2019sound}. 
Across set-ups, non-arbitrary preferences are found robustly across varying cultures and writing systems \cite{cwiek2022bouba}. Remarkably, to some extent this can even be found in blind individuals who undergo a haptic version of the bouba-kiki task \cite{bottini2019sound}.

\begin{table*}[ht]
    \centering
    \begin{tabular}{llllll}
        \textbf{Model}  & \textbf{Train objective}  & \textbf{Architecture} & \textbf{Attention}    & \textbf{\#Params} & \textbf{\#imgs,\#caps (M)} \\ \hline
         CLIP           & CON                       & Dual-Stream           & Modality-specific     & 151.3M            & 400, 400        \\
         ViLT           & ITM\&MLM                  & single-stream         & Merged                & 87.4M             & 4.10, 9.85   \\
         BLIP2          & CON\&IGTG\&ITM            & Dual-stream           & Q-Former              & $\sim$3.8B        & 129, 258              \\
         GPT-4o         & Unknown                       & Unknown                   & Unknown                   & Unknown               & Unknown           \\
    \end{tabular}
    \caption{Overview of the models. Objectives are Image Text Matching(ITM), Masked Language Modelling(MLM), Image-grounded Text Generation(IGTG), or Contrastive Learning(CON). Numbers are millions(M) or billions(B).}
    \label{tab:models}
\end{table*}

Most experiments in this domain are conducted using a two-alternative forced choice design, where two contrasting images are shown side by side (one jagged and the other curved), and two possible labels are offered, asking participants to make the "correct" mapping. However, it has been argued that this is an anti-conservative method in the sense that the concurrent presentation of two images that differ along one dimension and two labels that also differ along one dimension strongly primes participants to match the two, noticing their similarities. \citet{nielsen2013parsing} therefore introduced a different method, in which images are presented independently, and participants are asked to generate novel pseudowords to match the images. Here, we adopt their approach as a stringent method for probing VLMs for the bouba-kiki effect.

\subsection{Vision-and-language models}
\label{VLMs}
Despite recent advances in multi-modal models \cite{Zhang2024Vision} using transformer architectures, they remain poorly understood and often show unwanted behaviors such as poor visio-compositional reasoning \cite{Thrush_2022_CVPR, diwan-etal-2022-winoground} or spatial reasoning skills \cite{kamath-etal-2023-whats}. 
In addition, in the visual question-answering domain it is a well-known problem that models often lack visual grounding and have trouble integrating textual and visual data \cite{goyal2017making, jabri2016revisiting, agrawal2018don}. This makes it perhaps even more puzzling that \citet{alper2024kiki} found strong evidence for a bouba-kiki effect in CLIP and Stable Diffusion: even if these models are able to extract sound-symbolic information in the absence of auditory data, they will likely struggle to actually associate that information with visual properties. 

Their approach involved generating two large sets of pseudowords, where one set was more likely associated with round shapes (examples: \textit{bodubo, gunogu, momomo}) and the other set would evoke associations with jagged shapes (examples: \textit{kitaki, hipehi, texete}). The CLIP embedding vector space was used to define a visual semantic dimension that best separates two sets of pre-selected adjectives (various synonyms of round and jagged). Within this space, pseudoword properties could reliably predict adjective type (round or jagged), and geometric properties associated with those adjectives could predict the category of pseudowords. With Stable Diffusion, novel images were generated based on pseudowords and analyzed by embedding them using CLIP and through human evaluation. Both methods revealed evidence for the presence of sound symbolic mappings in these models \cite{alper2024kiki}.  

While their methods mainly involved text-to-image generation (with Stable Diffusion) or text-to-text mapping (with CLIP embeddings), we focus on image-to-text classification. We use images previously used in experiments with humans, as well as novel images generated following a procedure previously used to generate items for human experimentation. This approach provides an additional way of testing for cross-modal associations in VLMs and yields data that can be more directly compared to human data from studies into the bouba-kiki effect. Moreover, \citet{alper2024kiki} did not explicitly compare different VLMs (Stable Diffusion also uses CLIP). However, it would not be surprising if properties relating to the architecture, for example, affect the presence of this effect since these properties directly determine how the modality gap is bridged. Previous findings suggest that dataset diversity and scale are the primary drivers of alignment to human representations \cite{Conwell2023whatcan, muttenthaler2023human}. We compare four models here, with different architectures, attention mechanisms, and training objectives. 

While many different architectures exist, they typically use single or dual-stream architectures. Either combining the inputs from two modalities and encoding them jointly (single-stream) or encoding them by two separate modality-specific encoders (dual-stream). Single-stream architectures typically use merged attention, where the language and visual input attend to both themselves and the other modality. Dual-stream architectures often use some form of cross-model attention, like co-attention and modality-specific attention, in addition to merged attention. Recently, \citet{Li2023BLIP-2} introduced a lightweight Querying Transformer (Q-Former) to bridge the modality gap between any arbitrary pre-trained frozen vision model and a language model, resulting in BLIP2. Frequently, image text matching and masked language modeling are used as learning objectives \cite[e.g., ViLT; ][]{Kim2021VILT}, but some methods use a contrastive learning objective (e.g., CLIP) or use image-grounded text generation loss (e.g., BLIP, BLIP2). The models used in this paper are shown in Table~\ref{tab:models}. They are different in the above aspects, allowing investigation into the effect of their designs and input data on the bouba-kiki effect. In addition, we include GPT-4o; even though no information is available for this model, its generative performance is unprecedented. 


\section{Methods}
To test for the presence of a bouba-kiki effect in VLMs we employ previously used as well as newly generated images (§\ref{sec:image_gen}) and use a method for constructing pseudowords (\ref{psuedo_gen}) that is directly borrowed from \citet{nielsen2013parsing}. Probing (§\ref{probing}) was used to obtain image-text scores and responses were analyzed in two ways (§\ref{analysis}). 

\subsection{Image selection and generation}
\label{sec:image_gen}
The original set of images used by \citet{kohler1929gestalt, kohler1947gestalt}, as shown in figure \ref{fig:kohler}, has been expanded in subsequent experiments. \citet{maurer2006shape} for example introduced additional line drawings and \citet{westbury2005implicit} used images with white shapes on a black background. Here we use the original pair and the two sets of four image pairs by \citet{maurer2006shape, westbury2005implicit}. In addition, we generated new random curved and jagged images using a method inspired by \citet{nielsen2013parsing}. We generated 10 uniformly distributed points within a circle with a radius of 1. These points were connected with either smooth curves or straight lines. For curved images, we generated curves that pass through the given points such that they form a closed path. 
Jagged images were generated by connecting the ordered points with straight lines, also forming a closed path. All images are displayed in Appendix~\ref{sec:images}.

\subsection{Pseudoword generation}
\label{psuedo_gen}
Following the experiment conducted by \citet{nielsen2013parsing} with human participants, we present the VLMs with a constrained set of syllables that can be used to construct novel pseudowords. Based on previously established cross-modal association patterns, \citet{nielsen2013parsing} selected sets of vowels and consonants that were expected to evoke a sense of correspondence with either jagged or curved visual shapes. We adopt exactly their set here, consisting of sonorant consonants M, N and L and rounded vowels OO, OH and AH, expected to match to curved shapes, and plosive consonants T, K and P and non-rounded vowels EE, AY and UH, expected to match to jagged shapes. Syllables were created by making consonant-vowel combinations. In total 36 different syllables (e.g., loo, nah, kee, puh) can be constructed in this way, with nine different versions of each syllable type: sonorant-rounded (S-R), plosive-rounded (P-R), sonorant-non-rounded (S-NR) and plosive-non-rounded (P-NR).

In addition to single syllables, we generated pseudowords by concatenating two syllables, as this was exactly the task human participants were asked to complete in the experiment \cite{nielsen2013parsing}.
However, since we are not primarily interested here in distinguishing the separate roles played by consonants versus vowels in the bouba-kiki effect, and \citet{nielsen2013parsing} demonstrated that both have an effect, we limit the set of possible syllables in two-syllable probing to combinations of S-R syllables and P-NR syllables.

\begin{figure*}[h!]
  \centering
  \includegraphics[width=0.95\linewidth]{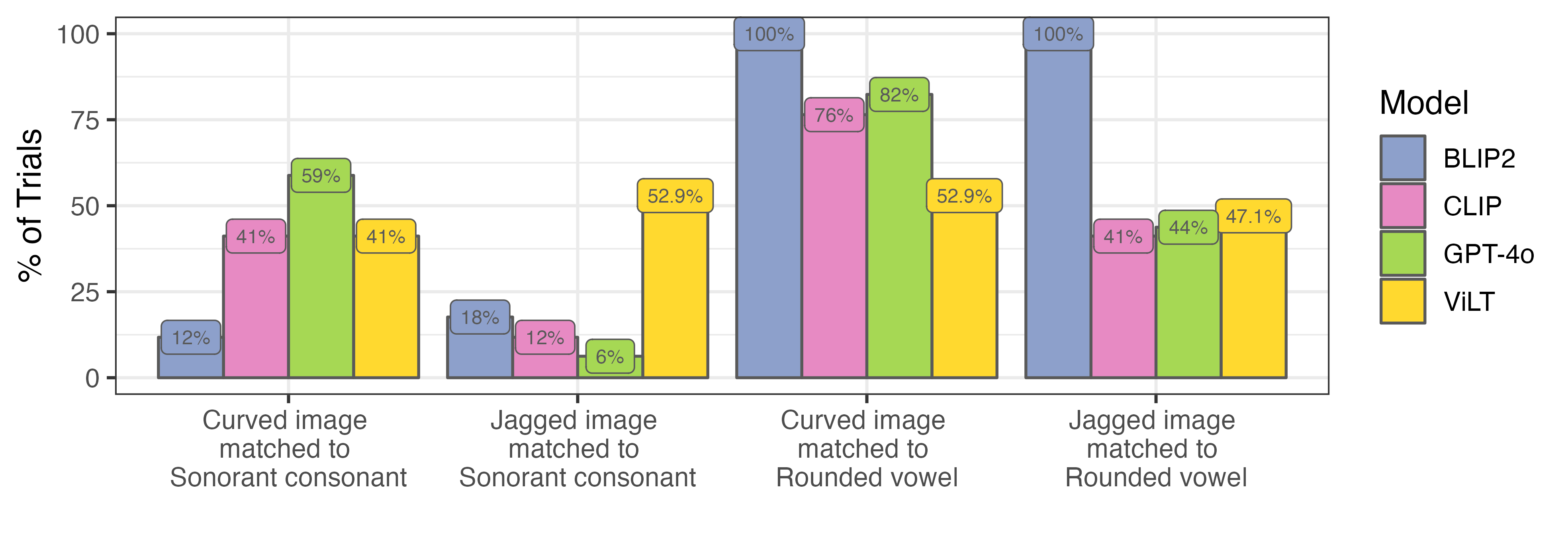}
  \caption{Percentages of trials in which selected syllables contain sonorant consonants or rounded vowels, separated by image shape (Jagged or Curved) for all four VLMs}
  \label{fig:single_syl}
\end{figure*}

An important difference between the human set-up and our work, is that their participants also listened to a spoken version of the pseudowords, while our models are only exposed to the written form. Since the bouba-kiki effect is most often assumed to integrate vision and sound, this may influence the result. However, the relation between orthographic shapes and the sounds they represent is not arbitrary either and has presumably been shaped by human iconic strategies in their development and evolution \cite{turoman2017glyph}. This perhaps also explains why a role for English orthography has been demonstrated in the bouba-kiki effect for humans \cite{cuskley2017phonological}, while at the same time it is robust across different writing systems \cite{cwiek2022bouba}.

\subsection{VLM probing}
\label{probing}
To assess the preferences of BLIP2, CLIP, and ViLT, in each trial, we extract probabilities for all possible labels (i.e., syllables and pseudowords) conditioned on an image. Instead of only embedding the label, each label is fed in a sentence (`The label for this image is \{label\}') such that embedding the textual input is closer to the models' natural objective\footnote{Additional analysis revealed that the overall results remain consistent even when only the label is provided.}. Importantly, only the labels differ between inferences such that variance in the probability given an image is caused by the label only.
Where \citet{alper2024kiki} use an \textit{indirect} metric by embedding the inputs in CLIP space, our method uses the model probabilities as a more \textit{direct} measure of how well a given syllable or pseudoword matches a novel image. For GPT-4o, we prompt the model to generate a label and use its probability directly (Appendix \ref{gpt-prompt}). 

\subsection{Analysis} 
\label{analysis}
All findings were analyzed for statistical significance using Bayesian models with the \textit{brms} package \cite{brms} in R \cite{r}. To analyze VLM probability scores, we fitted Bayesian multilevel linear models (4 chains of 4000 iterations and a warmup of 2000, family = \textit{gaussian}) to predict probability with image shape (Jagged versus Curved), consonant (plosive or sonorant) and vowel (rounded or non-rounded) categories ($Probability \sim shape * (consonant + vowel)$). For all models of this type, the random effects structure consists of varying intercepts for image and label with by-label random slopes for shape. When comparing proportions of vowels, consonants, or selected pseudoword types, we fitted Bayesian logistic models (4 chains of 1000 iterations and a warmup of 500, family = \textit{binomial}) to test whether shape predicts the occurrence of particular vowels, consonants or pseudoword types ($Occurrence | trials(Sample Size) \sim Shape$). Effects are considered significant when the computed 95\% Credible Interval does not include 0, i.e. the lower and the upper bounds of the CI have to be either both positive or both negative. All plots were created in ggplot2 \cite{wickham2016}.

\section{Results}
The findings are analyzed in two ways. First, we compare the results of VLM probing to the performance of human participants \cite{nielsen2013parsing}. For BLIP2, CLIP and ViLT this means we first only consider the syllable or pseudoword with the highest probability for each image. These are then analyzed similarly to those selected by humans or generated by GPT-4o. Second, we examine the probabilities for \textit{each} possible syllable or pseudoword from BLIP2, CLIP and ViLT, to obtain a more comprehensive measure of cross-modal associations. For the GPT-4o results reported below, one image in the Jagged shape condition is consistently missing since it (top right image in Figure \ref{fig:westbury} in Appendix~\ref{sec:images}) was flagged as `content that is not allowed by our safety system'.

\subsection{Single syllable selection} \label{subsection:syl}
VLMs were first probed using single syllables, here we are interested to see if the models predominantly pair Jagged images with P-NR and Curved images with S-R syllables, as was found with humans. 
Figure~\ref{fig:single_syl} shows these results as the percentage of trials (where each individual image of the set of 17 pairs forms a trial) in which model probabilities where highest for sonorant consonants or rounded vowels with either Curved or Jagged shapes.
A result that fits the expected human pattern would show higher bars for the Curved than for the Jagged shapes in both sets. The only models where this seems to go in the right direction are CLIP and GPT-4o. BLIP2 mostly displays a general preference for P-R syllables, without considering the shape and ViLT does not display any clear preference. To test whether the differences in percentages for CLIP and GPT-4o are significant, we use Bayesian logistic models (as described in \ref{analysis}). 
For both models, Jagged images are paired with sonorant consonants significantly less often than Curved images (CLIP: $b$ = -1.79, Bayesian 95 \% Credible Interval $[-3.86, -0.05]$, GPT-4o: $b$ = -3.51, 95 \% CI $[-6.69, -1.37]$) and Jagged images are paired with rounded vowels significantly less often than Curved images (CLIP: $b$ = -1.62, 95 \% CI $[-3.06, -0.19]$, GPT-4o: $b$ = -1.97, 95 \% CI $[-3.66, -0.36]$).

\begin{figure*}[h]
    \centering
    \includegraphics[width=\linewidth]{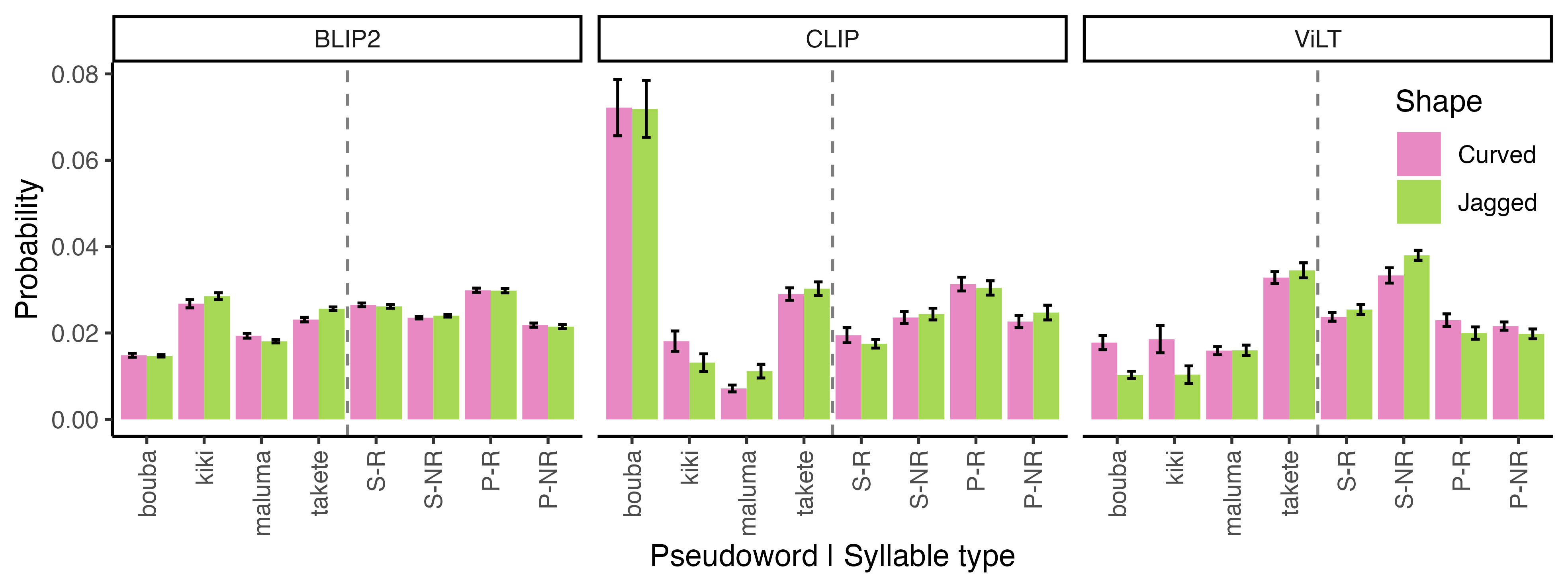}
    \caption{Probability scores for the original pseudowords (bouba, kiki, takete and maluma), as well as for the four different generated syllable types: Sonorant-Rounded (S-R), Sonorant-Non-Rounded (S-NR), Plosive-Rounded (P-R) and Plosive-Non-Rounded (P-NR), paired with two types of shapes (Jagged or Curved) for three VLMs}
    \label{fig:single_syl_probs}
\end{figure*}

\subsection{Probability scores for novel syllables}
While GPT-4o only selects the best fitting syllable out of all options for each image, CLIP, BLIP2 and ViLT provide probability scores for each possible syllable, yielding more comprehensive data. Here we therefore also analyze the probability scores for these three models, to investigate whether higher scores occur when pairing S-R syllables with Curved images than with Jagged images and vice versa for P-NR syllables. 
Figure \ref{fig:single_syl_probs} shows the probabilities for the pseudoword pairs that were used in the classic experiments with humans (bouba \& kiki, takete \& maluma) and the four different syllable types (S-R, S-NR, P-R, P-NR). 

Looking at the original pseudowords, none of the models display a clear bouba-kiki or takete-maluma effect. Probabilities for the different words differ overall (with a curiously high probability for "bouba" in CLIP), but this does not seem modulated by the visual shape. For the syllables, BLIP2 shows no shape-modulated variation at all, and ViLT displays contradictory patterns (e.g. higher probability scores for S-NR than S-R syllables with Curved shapes and higher scores for S-NR with Jagged than with both P-R and P-NR). Only CLIP gets close to the expected pattern, with equal scores for the ambiguous syllable types (S-NR and P-R) but slightly higher scores for P-NR with Jagged and S-R with Curved. Yet, no significant effects are found when testing whether CLIP shows a pattern of preferring the expected consonants and vowels with their associated shapes using a Bayesian multilevel linear model (as described in \ref{analysis}). For ViLT, we find one (tiny) interaction between shape and consonants in the opposite direction of what is expected, where scores for Jagged shapes are significantly higher when paired with sonorant versus plosive consonants ($b$ = .0056, 95 \% CI $[.0001, .0112]$). For BLIP2, we find a significant overall preference for rounded vowels ($b$ = 0.0055, 95 \% CI $[.0019, .0091]$), but no other effects.

\subsection{Two-syllable pseudoword selection}
\begin{figure}[t]
    \centering
    \includegraphics[width=\columnwidth]{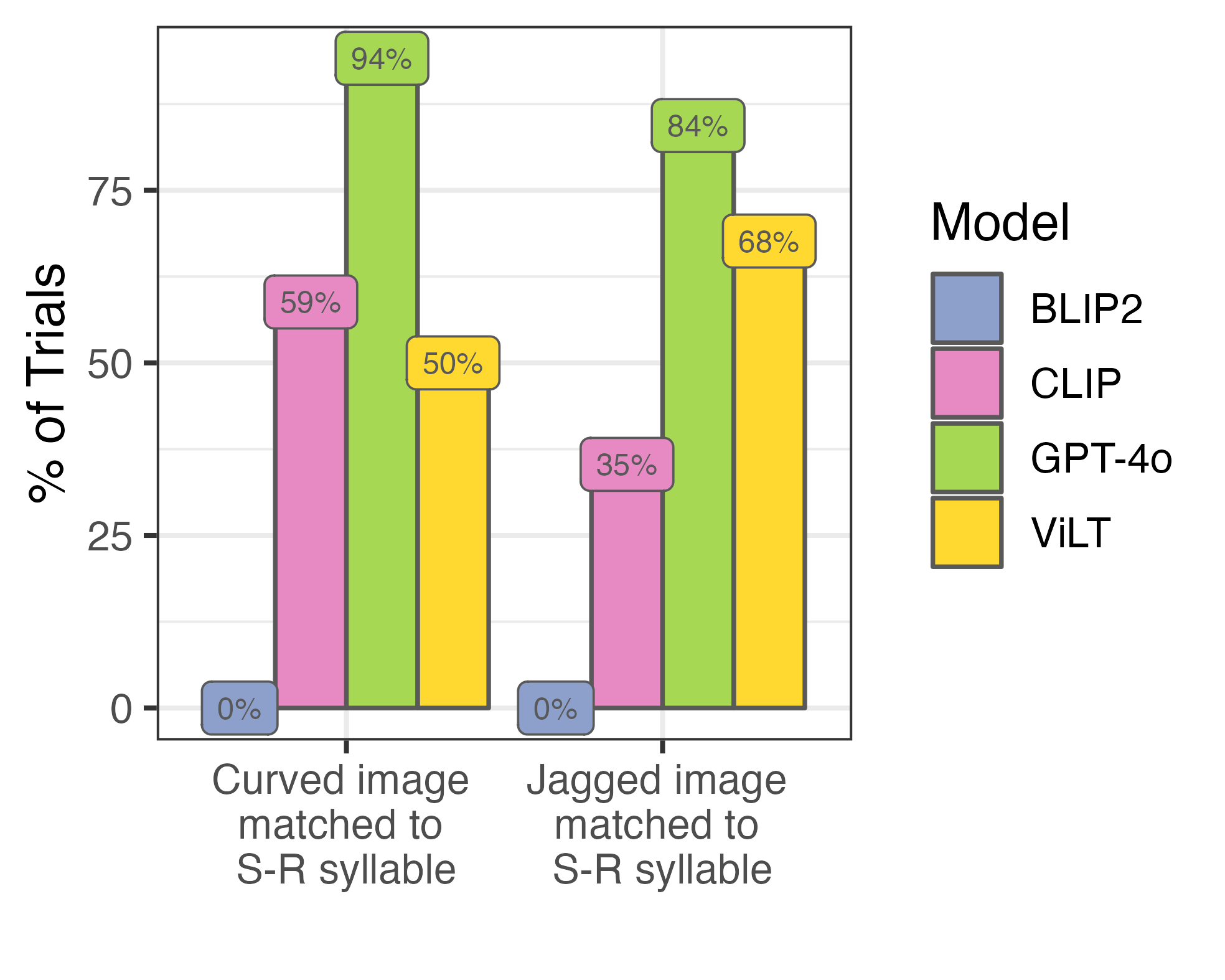}
    \caption{Percentages of trials in which Jagged or Curved visual shapes were matched to Sonorant-Rounded (S-R) syllables embedded in two-syllable pseudowords for all VLMs. Here 0\% for S-R syllables implies a 100\% preference for P-NR syllables.}
    \label{fig:two_syl}
\end{figure}

Although the results in \citet{nielsen2013parsing} were analyzed by looking at single syllables, the actual task human participants performed involved creating novel pseudowords consisting of two syllables. We therefore also used our VLMs to generate (GPT-4o) or provide probability scores (CLIP, BLIP2 and ViLT) for two-syllable pseudowords that were created by concatenating two of the possible syllables from the set of S-R (most Curved) and P-NR (most Jagged) syllables resulting in 324 words. For CLIP, BLIP2 and ViLT we first look at the "preferred" pseudowords, by only considering the option with the highest probability score for each image. Figure \ref{fig:two_syl} shows the percentages of trials in which S-R syllables were matched to either Curved or Jagged images, counting each one of the two syllables in a word separately. BLIP2 never used S-R syllables and only selected pseudowords that contained two P-NR syllables, independently from which image was shown. Both CLIP and GPT-4o show a higher percentage of Curved matched to S-R compared to Jagged, but GPT-4o seems to mostly just prefer S-R syllables overall. A manual inspection of GPT-4o's generated pseudowords revealed that in 25 out of 33 trials the word "nohmoh" was used, 12 times for Jagged and 13 times for Curved images. For ViLT, if a preference is present, it is in the wrong direction. In the case of CLIP, we find that Jagged images are indeed paired with S-R syllables significantly less often than Curved images ($b$ = -1.00, 95 \% CI $[-2.04, -0.04]$).

\subsection{Probability scores for novel two-syllable pseudowords}
\label{pseudo_probs}
\begin{figure*}[h]
    \centering
    \includegraphics[width=0.9\linewidth]{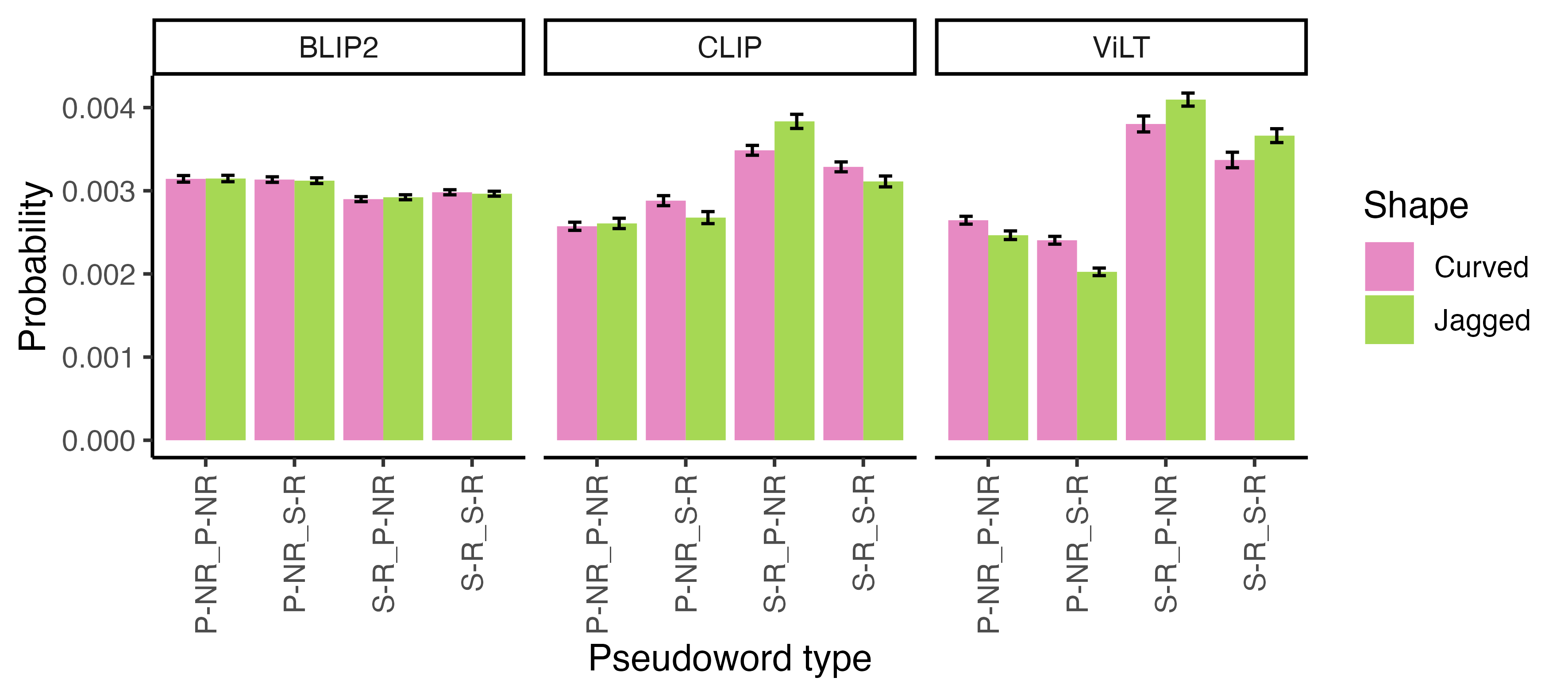}
    \caption{Probability scores for four pseudoword types, combining Sonorant-Rounded (S-R) and Plosive-Non-Rounded (P-NR) syllables, paired with two types of shapes (Jagged or Curved) for three VLMs}
    \label{fig:two_syl_probs}
\end{figure*}

We obtained probability scores for all possible two-syllable pseudowords when paired with each image for CLIP, BLIP2 and ViLT. Figure \ref{fig:two_syl_probs} shows these results by plotting probabilities for four different pseudoword types. The pseudoword on the left combines two P-NR syllables and is therefore expected to result in higher probabilities for Jagged shapes. Conversely, the most right pseudoword combines two S-R syllables and should evoke higher probabilities for Curved shapes. A pattern in which pink (Curved) bars rise while green (Jagged) bars fall would therefore reflect evidence for the bouba-kiki effect. None of the tested VLMs fit this pattern. Since GPT-4o generated "nohmoh" (and similar variants like "moomoh") almost exclusively when given the freedom to select two syllables from the full set of Jagged-associated and Curved-associated syllables, we also independently obtained probabilities for both syllable types. For this, we asked GPT-4o to generate a pseudoword for each image twice, once when given only the set of Jagged-associated syllable options, and once with only the Curved-associated syllables as options. Yet, again no significant effect of shape on probability scores for different syllable types was found. Figure \ref{fig:gpt_probs} in Appendix \ref{sec:gpt_probs} shows this result.

\subsection{Summary}
In summary, the bouba-kiki effect appeared absent for BLIP2 and ViLT, while for CLIP and GPT-4o, the results varied depending on how the effect was tested, and results were analyzed. When asking the model to select one best-fitting syllable, CLIP and GPT-4o both display the effect in the expected direction. However, this pattern disappears when looking at a richer dataset of probability scores (from CLIP, BLIP2, and ViLT) for each possible syllable. In the case of two-syllable words, GPT-4o results no longer display significant evidence for a bouba-kiki effect. 

\section{Discussion}

Our findings partly contradict previous work, which found that sound-symbolic associations are present in CLIP and Stable Diffusion \cite{alper2024kiki}. We use a different method, focusing on image-to-text probabilities, which is more similar to how the effect has been tested with humans. We show that it is too early to conclude that VLMs understand sound-symbolism or map visio-linguistic representations in a human-like way since the results depend heavily on which specific model is tested and how the task is formulated. This is unsurprising given that CNN-based models often classify based on superficial textural rather than shape features \cite{baker2018deep, geirhos2018imagenettrained, herman2020originsbias} and, albeit less so, this texture bias is also present in vision transformers \cite{geirhos2021partial}. Moreover, \citet{darcet2024vision} identified that, during inference, ViT networks create artifacts at low-informative background areas of images that are used for computations rather than describing visual information. Both findings are in stark contrast with what, at its core, is required for sound symbolism. However, the fact that some evidence for a bouba-kiki effect could be found in two of the four models tentatively suggests that real-world physical experience with different object properties may not be needed to develop this cross-modal preference but that it can, to some extent, be learned from statistical regularities in data containing text and images. 

Human language on its own already contains many non-arbitrary regularities between speech sounds and meaning \cite{blasi2016sound}, and these regularities, like phonesthemes \cite{bergen2004psychological}, can be detected and interpreted by models such as word embeddings \cite{abramova-fernandez-2016-questioning} and LSTM based language models \cite{pimentel-etal-2019-meaning}. No visual input is needed for this, and perhaps this is also what caused the appearance of the bouba-kiki effect in the work by \citet{alper2024kiki}. In our work, we gave more prominence to the visual input and found much less convincing evidence for the effect.

Regarding the design features of the models we tested, we see that the model with the best bouba-kiki alignment to human preferences, CLIP, is also trained on the largest amount of data (comparing the three models we have information on, not including GPT-4o). This finding aligns with previous work showing that dataset properties affect alignment with human representations \cite{Conwell2023whatcan, muttenthaler2023human}. However, despite having much more parameters than CLIP, BLIP2 does not show the effect. In addition, while both BLIP2 and CLIP use dual-stream architectures, only CLIP, which uses modality-specific attention mechanisms, displays some evidence of a bouba-kiki effect. Despite impressive performance on vision-language tasks, the Q-Former in BLIP2 apparently does not promote sound-symbolic associations. This is important knowledge for developing models with vision-language representations that align with those of humans. More aligned models show more robust few-shot learning \cite{sucholutsky2023alignment} and promote more natural interactions between humans and machines \cite{kouwenhoven2022emerging}. Although we find modest evidence for a bouba-kiki effect in GPT-4o, we cannot know the origin of this effect as model details are unknown.

\section{Conclusion}
Given the pervasive role cross-modal associations play in human linguistic processing, learning and evolution, we tested for the presence of a bouba-kiki effect in four VLMs that differ along various dimensions such as architecture design, training objective, number of parameters, and input data. Evidence for this effect is limited, but not entirely absent, in the tested VLMs and these findings inform discussions on the origins of the bouba-kiki effect in human cognition and future developments of VLMs that align well with human cross-modal associations.

\section{Limitations}
Our work has a few notable limitations. First, we used synthetic images that were previously used in experiments with humans. Even though this makes our results easily comparable to those of human studies, there is a potential risk that these images are out-of-domain for models that are predominantly trained on realistic images. In future extensions of this work we therefore plan to include more naturalistic images. 

A second limitation manifests itself in the tokenization of the textual input. While humans in the experiment evaluate pseudowords as a whole, the tokenization process in language models may split our syllables or pseudowords into tokens that would not necessarily evoke the expected cross-modal associations in humans either (e.g., a separate evaluation of H in OH may invite a jagged association instead of curved). Despite being a fundamental difference, the primary goal of this work was to assess the preferences of VLMs in their most basic form. Further work should investigate whether tokenization affects results and identify whether there may be model-specific cross-modal associations on a token instead of word level. 

Third, the pseudowords we used were based on an experiment with humans but were different from those used by \citet{alper2024kiki}, who did find a strong bouba-kiki effect in CLIP embeddings. To allow for a better comparison with their findings, future work should also test our image-to-text approach with their set of pseudowords.

Finally, our experiments included a relatively small number of trials, limited by available experimental stimuli from human studies. By combining images from several previous studies and augmenting this set with additional newly generated images, we used more trials than most studies conducted with humans, though. The set of generated images can easily be expanded in future work. However, given the current pattern of results, this is not expected to lead to a more robust bouba-kiki effect in most models.


\bibliography{anthology, custom}

\appendix

\section{Full set of images}
\label{sec:images}

\begin{figure}[t]
   \includegraphics[width=0.48\linewidth]{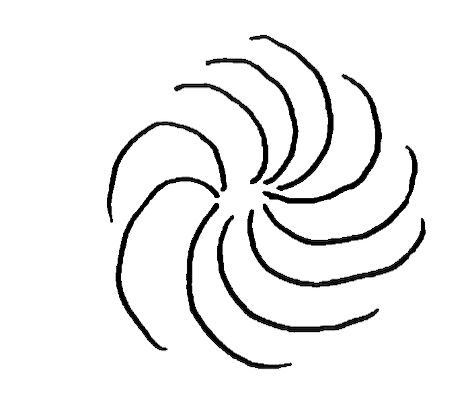}  \hfill
  \includegraphics[width=0.48\linewidth]{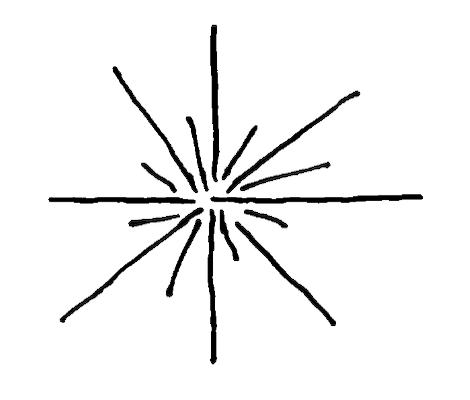}
   \includegraphics[width=0.48\linewidth]{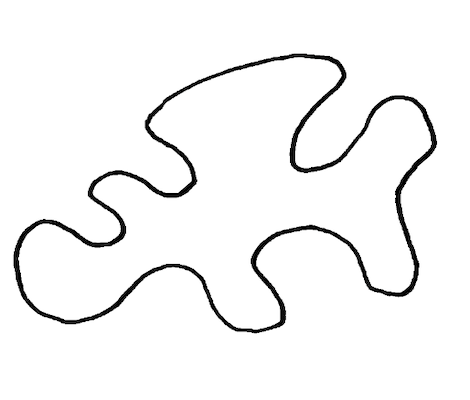}  \hfill
  \includegraphics[width=0.48\linewidth]{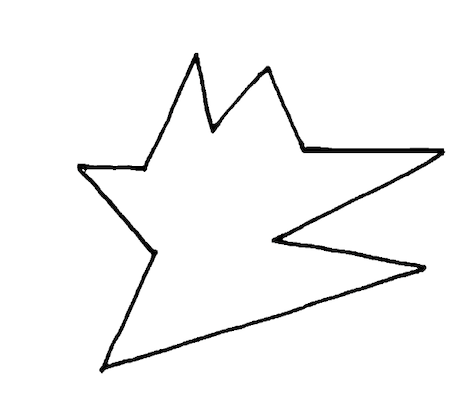}
   \includegraphics[width=0.48\linewidth]{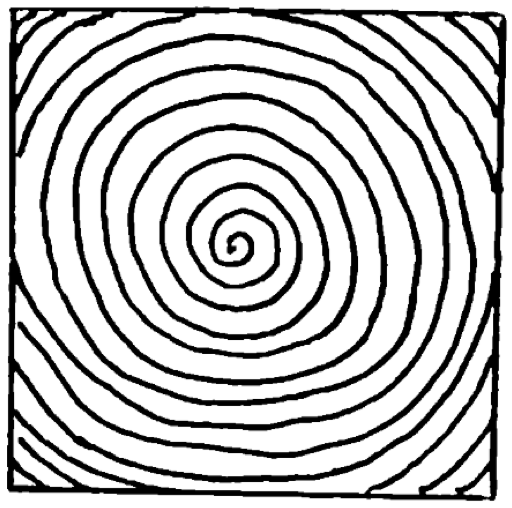}  \hfill
  \includegraphics[width=0.48\linewidth]{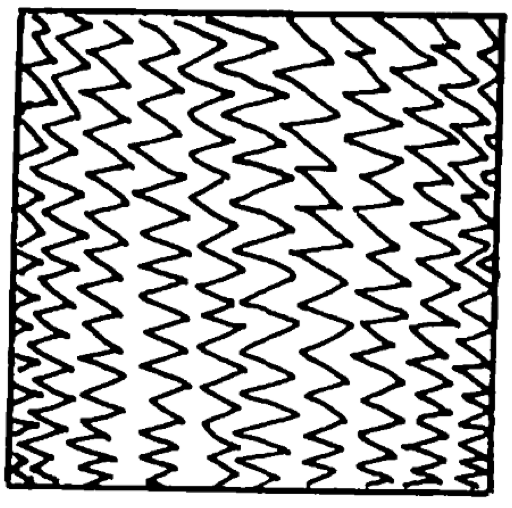}
   \includegraphics[width=0.48\linewidth]{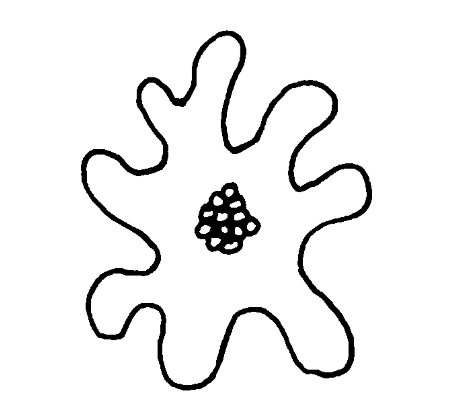}  \hfill
  \includegraphics[width=0.48\linewidth]{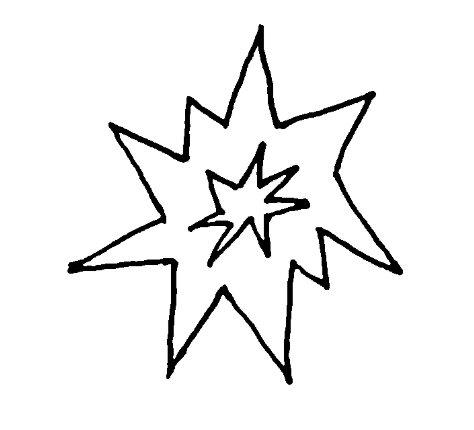}
  \caption{Images from \citep{maurer2006shape}}
  \label{fig:maurer}
\end{figure}

\begin{figure}[t]
   \includegraphics[width=0.48\linewidth]{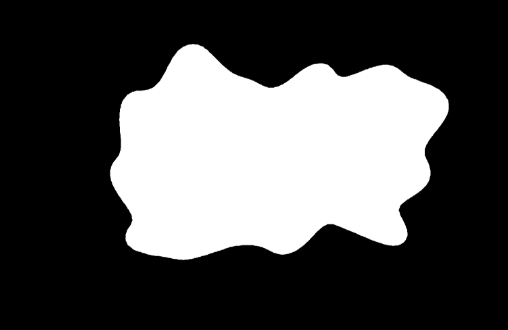}  \hfill
  \includegraphics[width=0.48\linewidth]{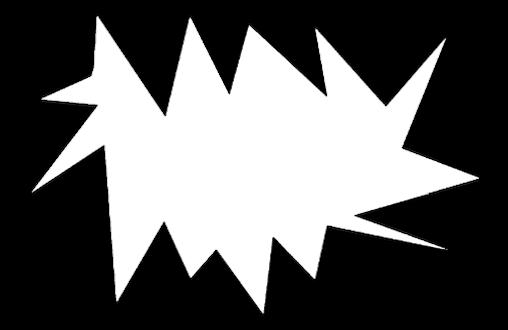}
   \includegraphics[width=0.48\linewidth]{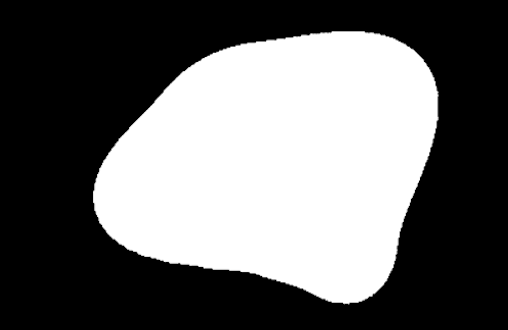}  \hfill
  \includegraphics[width=0.48\linewidth]{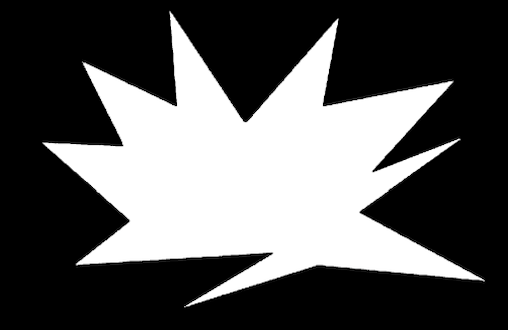}
   \includegraphics[width=0.48\linewidth]{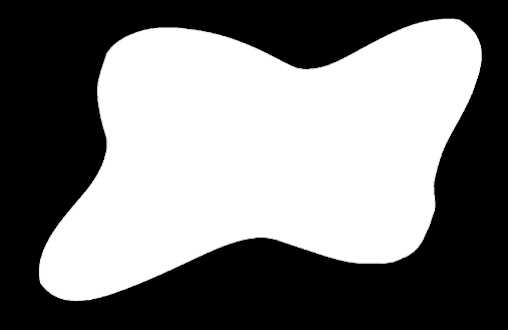}  \hfill
  \includegraphics[width=0.48\linewidth]{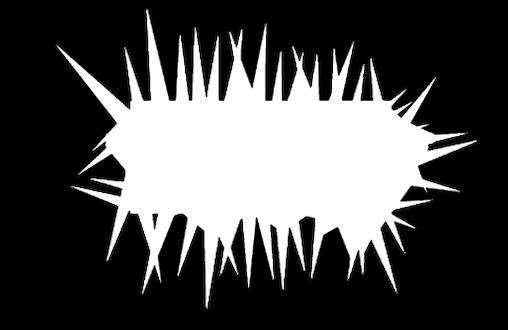}
   \includegraphics[width=0.48\linewidth]{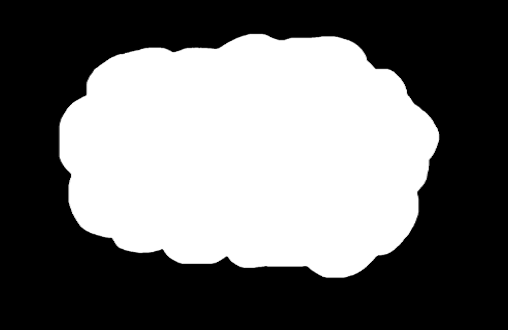}  \hfill
  \includegraphics[width=0.48\linewidth]{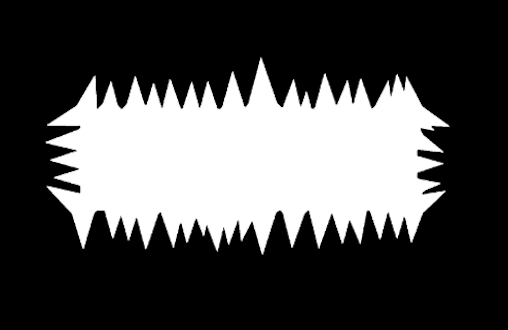}
  \caption{Images from \citep{westbury2005implicit}}
  \label{fig:westbury}
\end{figure}

\begin{figure}[t]
   \includegraphics[width=0.4\linewidth]{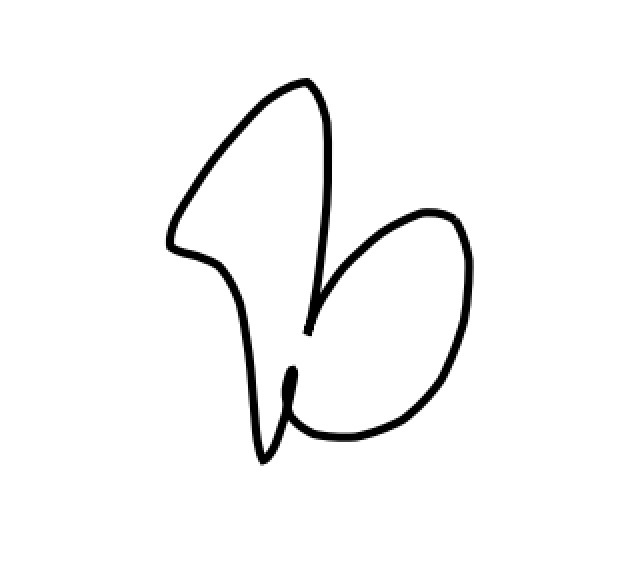}  \hfill
  \includegraphics[width=0.4\linewidth]{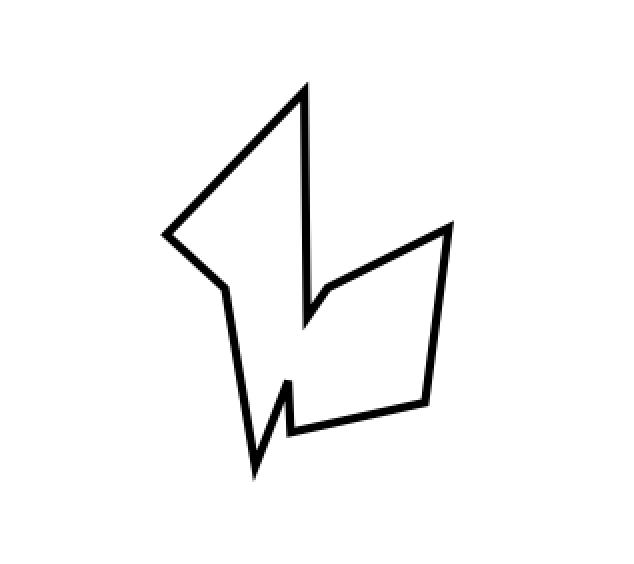}
   \includegraphics[width=0.4\linewidth]{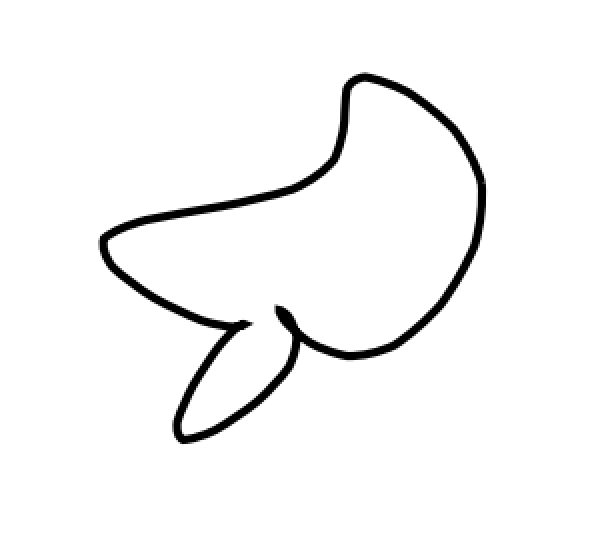}  \hfill
  \includegraphics[width=0.4\linewidth]{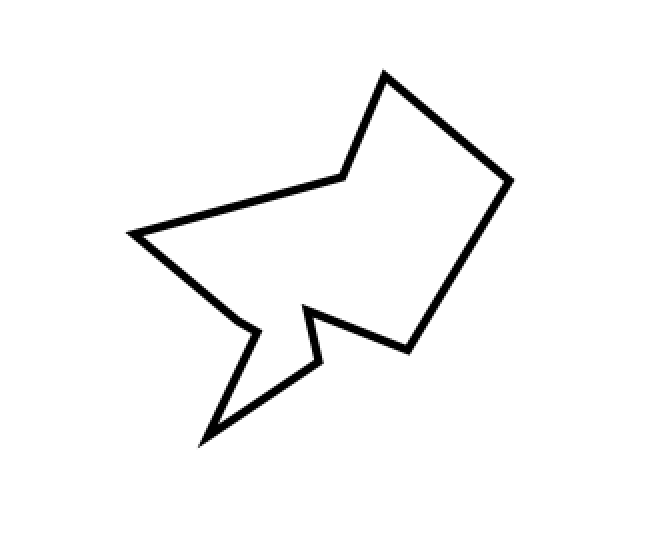}
   \includegraphics[width=0.4\linewidth]{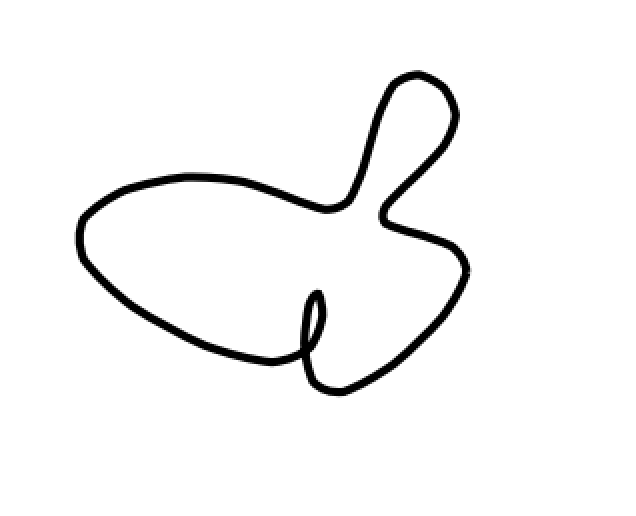}  \hfill
  \includegraphics[width=0.4\linewidth]{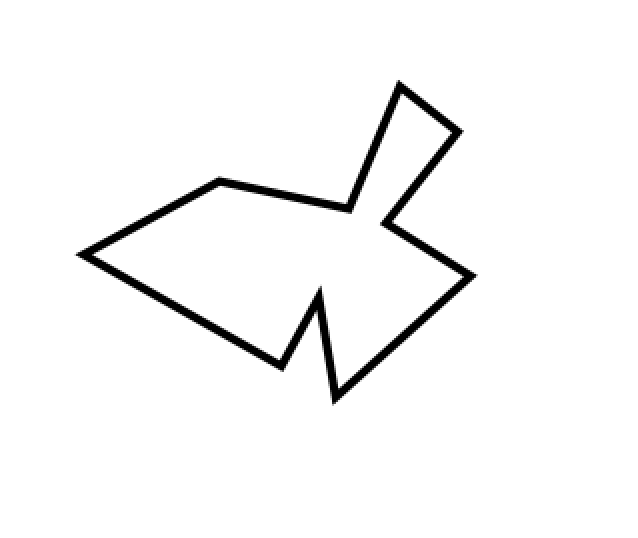}
   \includegraphics[width=0.4\linewidth]{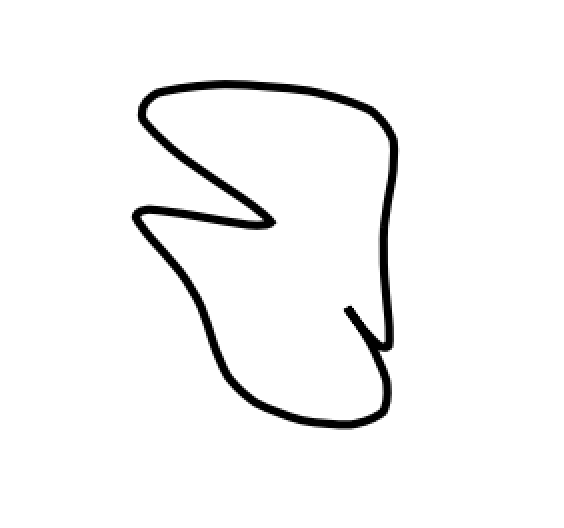}  \hfill
  \includegraphics[width=0.4\linewidth]{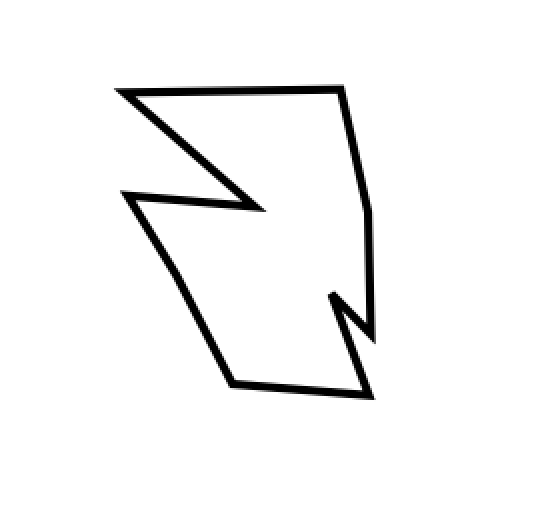}
\includegraphics[width=0.4\linewidth]{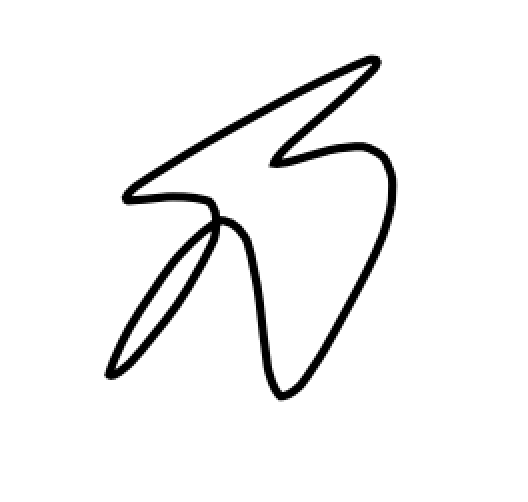}  \hfill
  \includegraphics[width=0.4\linewidth]{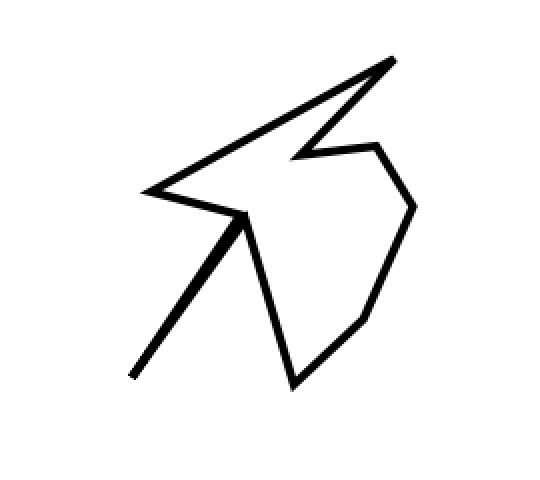}
   \includegraphics[width=0.4\linewidth]{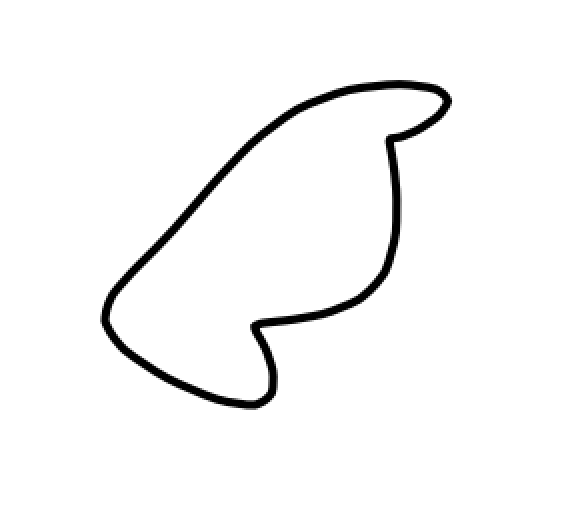}  \hfill
  \includegraphics[width=0.4\linewidth]{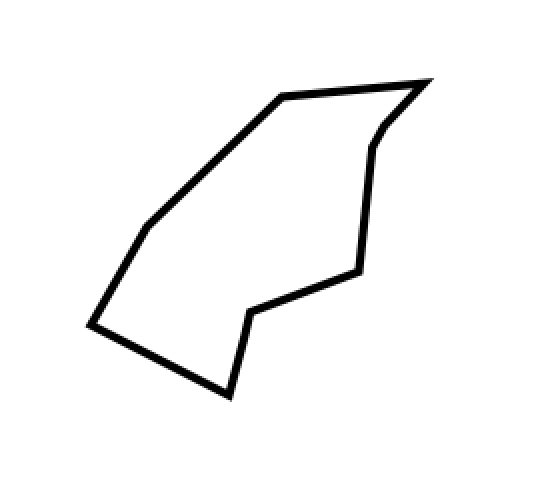}
   \includegraphics[width=0.4\linewidth]{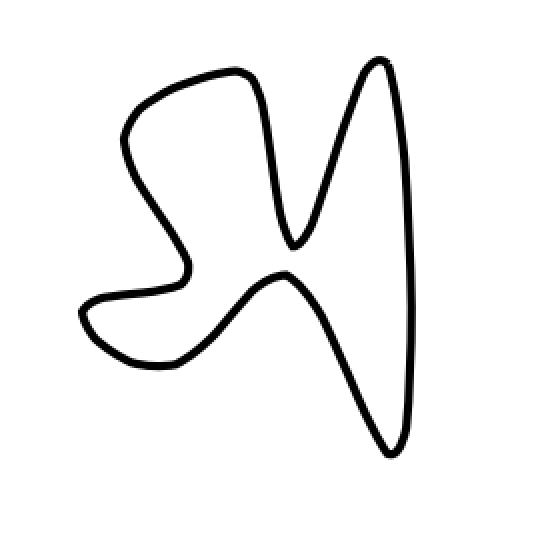}  \hfill
  \includegraphics[width=0.4\linewidth]{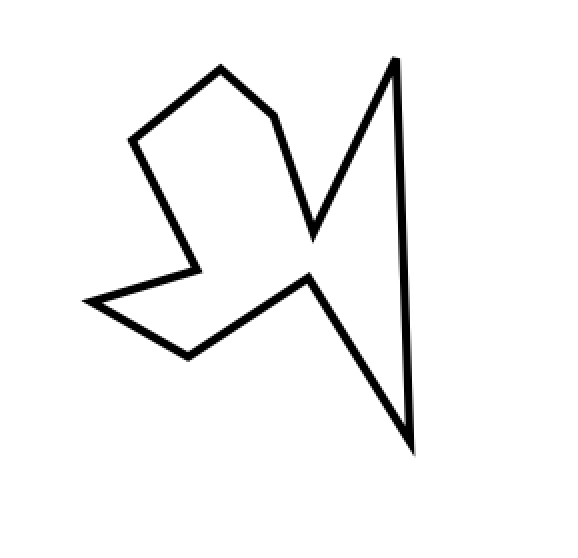}
   \includegraphics[width=0.4\linewidth]{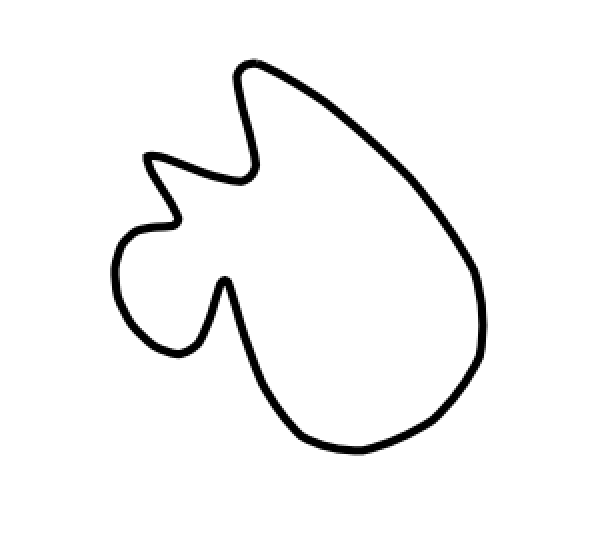}  \hfill
  \includegraphics[width=0.4\linewidth]{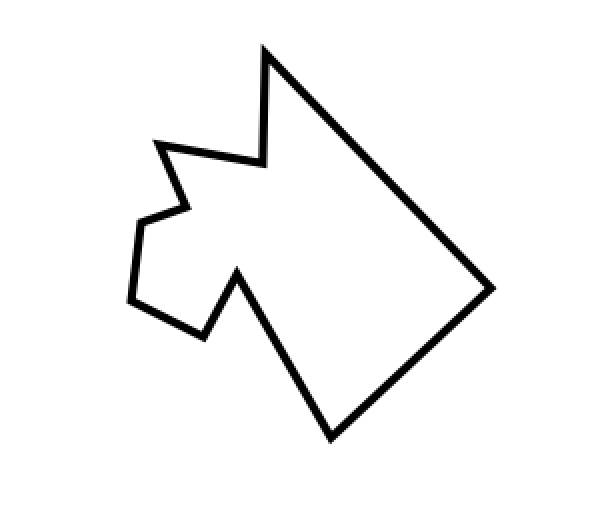}
  
  \caption{Newly generated images}
  \label{fig:generated}
\end{figure}

This appendix presents the full set of images with visual shapes that were used in the experiments. Besides the original image pair from \citet{kohler1929gestalt, kohler1947gestalt} which was shown in Figure~\ref{fig:kohler}, we used four image pairs from \citet{maurer2006shape}, displayed in figure~\ref{fig:maurer}, four from \citet{westbury2005implicit}, displayed in figure \ref{fig:westbury}, and 8 additional pairs we newly generated using a method inspired by the one described by \citet{nielsen2013parsing}, displayed in Figure~\ref{fig:generated}. For each image pair, the Curved version is displayed on the left and the Jagged version on the right.

\section{GPT-4o prompting}
\label{gpt-prompt}
Image-label matching is not directly possible for GPT-4o since the probabilities of the input tokens cannot be accessed. We therefore prompt (\ref{fig:prompt}) this model, with the temperature being 0.0, to generate a syllable or pseudoword given an image and use the log probabilities of the generated tokens to calculate the probability for a label conditioned on an image. Just like in the sentence setup used in the other models, our interest lies not primarily in the variability that may arise from using different prompts but rather focuses on the influence of the image on the predictions by using a simple and effective prompt that is identical for each image. Doing so allows us to use the resulting probabilities as a gauge for the models' preference of a label for a given image. 

\begin{prompt}[!ht]
\small
\begin{lstlisting}
You are given an image for which you need to assign a label. Use {one/two} of the following labels: {possible_labels}. Only respond with the label.
\end{lstlisting}
\caption{The exact prompt used to obtain GPT-4o probabilities. $possible\_labels$ corresponds to the syllables of interest.}
\label{fig:prompt}
\end{prompt}

\section{GPT-4o pseudoword probabilities}
\label{sec:gpt_probs}

\begin{figure}[t]
  \includegraphics[width=\columnwidth]{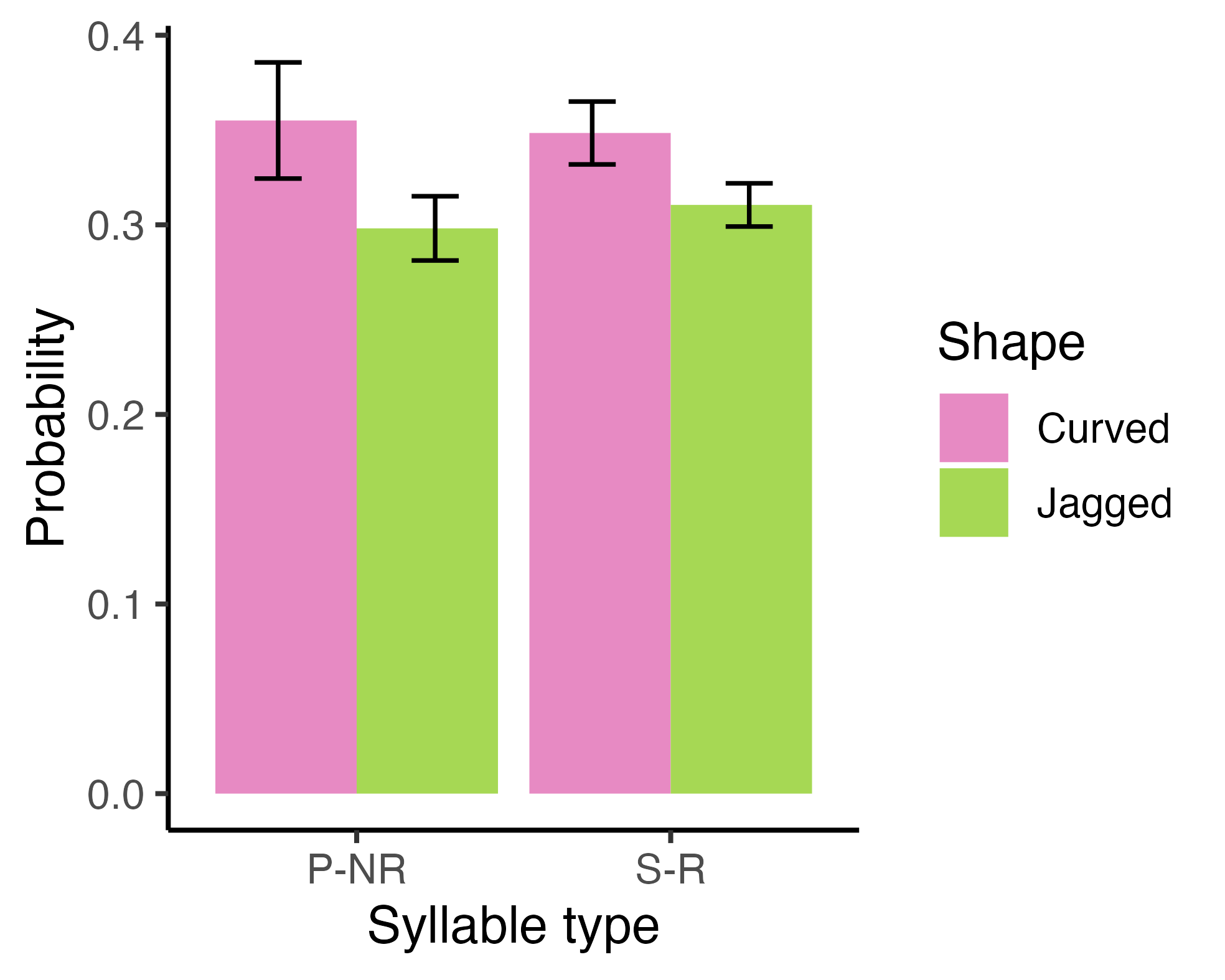}
  \caption{Probability scores for GPT-4o when forced to generate a pseudoword for each image twice, once by combining two Jagged-associated syllables, and once with only the Curved-associated syllables as options.}
  \label{fig:gpt_probs}
\end{figure}

In section \ref{pseudo_probs} we describe results for an experiment in which we asked GPT-4o to generate a pseudoword for each image twice, once when given only the set of Jagged-associated syllable options, and once with only the Curved-associated syllables as options. Figure \ref{fig:gpt_probs} shows the probabilities associated with these generated pseudowords. As concluded in the main text, no evidence for a preference to match P-NR syllables with Jagged shapes and S-R syllables with Curved shapes was found.

\end{document}